\documentclass[letterpaper, 10 pt, conference]{ieeeconf}  

\IEEEoverridecommandlockouts                              

\usepackage{graphicx} 
\usepackage{amsmath} 
\usepackage{amssymb}  

\usepackage{xcolor}

\title{\LARGE \bf
Learning Quadruped Locomotion using Bio-Inspired Neural Networks with Intrinsic Rhythmicity
}

\author{Chuanyu Yang$^{1*}$, Can Pu$^{1*\dagger}$, Tianqi Wei$^{2}$, Cong Wang$^{3}$, Zhibin Li$^{4}$
\thanks{This research is supported by Shenzhen Amigaga Technology Co. Ltd., by the Human Resources and Social Security Administration of Shenzhen Municipality under Overseas High-Caliber Personnel project (Grant NO. 202102222X, Grant NO. 202107124X) and by Human Resources Bureau of Shenzhen Baoan District under High-Level Talents in Shenzhen Baoan project (Grant No. 20210400X, Grant No. 20210402X).}
\thanks{* These authors contributed equally to this work.}
\thanks{$\dagger$ Corresponding author. Email: can.pu@amigaga.com}
\thanks{$^{1}$Chuanyu Yang and Can Pu are with Shenzhen Amigaga Technology Co Ltd. {$\{$chuanyu.yang, can.pu$\}$@amigaga.com}}%
\thanks{
$^{2}$Tianqi Wei is with the School of Artificial Intelligence, Sun Yat-sen University, CN.}
\thanks{
$^{3}$Cong Wang is with Shenyang Institute of Automation, Chinese Academy of Sciences, CN.}
\thanks{
$^{4}$Zhibin Li is with the Department of Computer Science, University College London, UK.}
}

\begin{document}

\maketitle
\thispagestyle{empty}
\pagestyle{empty}

\begin{abstract}
Biological studies reveal that neural circuits located at the spinal cord called central pattern generator (CPG) oscillates and generates rhythmic signals, which are the underlying mechanism responsible for rhythmic locomotion behaviors of animals. Inspired by CPG's capability to naturally generate rhythmic patterns, researchers have attempted to create mathematical models of CPG and utilize them for the locomotion of legged robots. In this paper, we propose a network architecture that incorporates CPGs for rhythmic pattern generation and a multi-layer perceptron (MLP) network for sensory feedback. We also proposed a method that reformulates CPGs into a fully-differentiable stateless network, allowing CPGs and MLP to be jointly trained with gradient-based learning. The results show that our proposed method learned agile and dynamic locomotion policies which are capable of blind traversal over uneven terrain and resist external pushes. Simulation results also show that the learned policies are capable of self-modulating step frequency and step length to adapt to the locomotion velocity.


\end{abstract}

\section{INTRODUCTION}

Animals are able to adapt their locomotion gait pattern to suit the locomotion velocity and ground condition. Efforts have been put into discovering the underlying mechanism of animal locomotion, and have obtained evidence that legged locomotion is rhythmic in nature. Findings have revealed evidence of special neurons called central pattern generators in the animal spinal cord. CPGs are a series of coupled oscillatory neurons capable of producing rhythmic signals internally without external sensory input. The rhythmic signals produced by CPGs are necessary for activities that involve periodic movement, such as breathing, walking, running, swimming, flying, etc. \cite{yu2013surveyOn}, \cite{ijspeert2008centralPattern}.

In recent years, there has been significant progress in using deep reinforcement learning (DRL) to train artificial neural networks for legged locomotion. Multi-expert neural networks can be designed to perform various different locomotion tasks \cite{yang2020multiExpert}. Robust control policy has also been trained to successfully traverse over extreme terrain in the wild \cite{lee2020learningQuadrupedal}. The neural networks from both works rely on an external periodic phase input to produce rhythmic motor patterns. Apart from a single global phase, multiple local phases can be provided, e.g. providing four phases as input to match the four individual legs of the quadruped \cite{shao2021learningFree}.

Instead of providing external periodic phase input, there have been attempts in designing network structures capable of self-generating phases. One approach is to add an additional phase increment output to the policy. The phase increment is added to the current phase and is used as the phase for the next timestep in a bootstrap manner \cite{sheng2022bioInspired}, \cite{yang2022fastAnd}, \cite{lee2020learningQuadrupedal}. To reproduce the agile locomotion behaviors of animals in legged robots, researchers have looked into animals for inspiration. Inspired by findings of CPGs within the animal nervous system, there have also been attempts in utilizing the intrinsic rhythmicity of CPGs to generate rhythmic motor patterns for locomotion controllers in robotics, eliminating the need for external phase input \cite{cho2019adaptationToEnvironmental}, \cite{nakamura2007reinforcementLearning}, \cite{fukunaga2004reinforcementLearning}.

\begin{figure}[t!]
  \centering
  \includegraphics[width=0.48\textwidth, trim={0 9cm 0 12cm},clip]{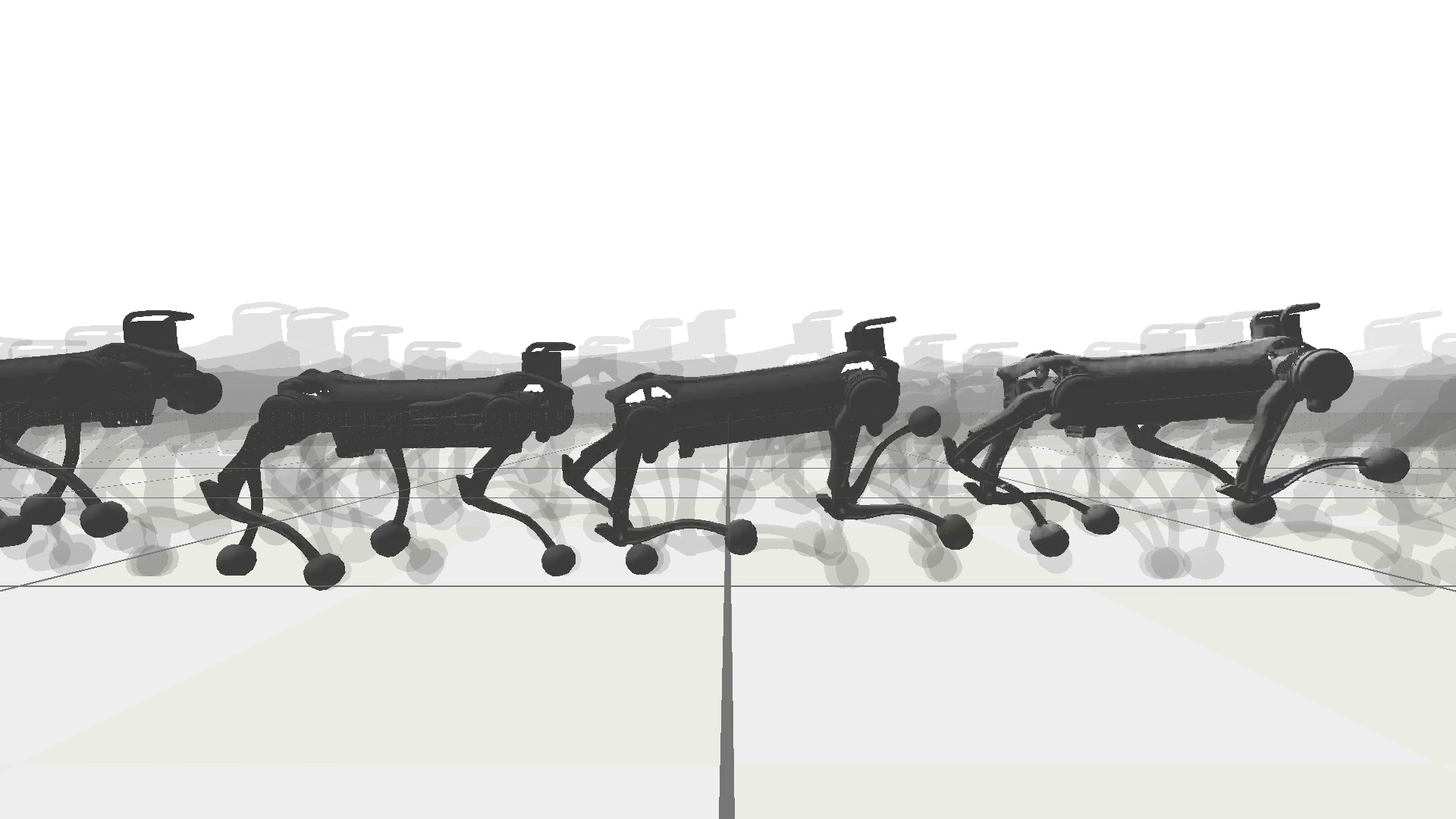}
  \caption{Depiction of the agile and dynamic locomotion learned by the MLP-CPG network architecture.}
  \label{side_overlay}
  \vspace{-4mm}
\end{figure}

This work aims to utilize the natural rhythmicity of CPGs to generate agile and dynamic locomotion behaviors for quadruped robots. While CPGs by themselves are capable of generating open-loop rhythmic signals, sensory feedback has to be integrated to enable the CPGs to react to external environment. We propose a framework that combines MLP networks and CPGs, where the CPGs are responsible for generating rhythmic patterns, and the MLP receives sensory information and provides feedback to modulate the CPGs' behavior. Our proposed framework reformulates CPGs by exposing the internal states as inputs and outputs of the network, essentially turning CPGs into a fully-differentiable stateless network. The parameters of the MLP and CPGs are optimized in conjunction via DRL. 


The contributions of this paper are:
\begin{itemize}
    \item We proposed a network architecture that combines CPGs and MLP network to generate rhythmic motions intrinsically without relying on external phase input. We name the network architecture as MLP-CPG.
    \item We proposed an approach that reformulates CPGs into fully-differentiable networks. The CPG parameters and MLP parameters within the MLP-CPG network architecture can be jointly optimized with DRL.
    \item We implemented the network architecture on a simulated quadruped robot and demonstrated the policy's capability to track user targets, and the robustness to environmental uncertainties.
    \item The proposed network architecture is interpretable. Locomotion characteristics such as step frequency, step length, and gait pattern can be inferred from the CPG parameters.
\end{itemize}

\section{RELATED WORK}
\subsection{CPG-Based Locomotion for Legged Robots}

CPGs have gained interest within the robotics field due to their oscillatory and rhythmic nature. To study the rhythmic mechanism of biological CPGs, researchers have proposed different CPG models with varying levels of complexity, from detailed biophysical models, connectionist models, to abstract models based on mathematical oscillators \cite{ijspeert2008centralPattern}, \cite{aoi2017adaptiveControl}.

Researchers have attempted to leverage the natural rhythmic behaviors of CPGs by incorporating CPGs into the controller to achieve animal-like locomotion for all kinds of legged robots, including bipeds, quadrupeds, and hexapods. Various abstract models of mathematical oscillators such as SO(2) oscillator, Matsuoka oscillator, and Hopf oscillator have been employed within the locomotion controller for different robots \cite{thor2019fastOnline}, \cite{endo2005experimentalStudies}, \cite{gay2013learningRobot}. 


\subsection{Learning CPG-Based Controllers with DRL}
DRL has proven to be effective in learning robust and complex locomotion skills for legged robots in simulation and real world. There are many works that have combined CPGs together with MLP networks and trained using the DRL paradigm. However, due to the intrinsic recurrent nature of CPG networks, back-propagation through time (BPTT) is required to obtain the parameters of CPGs while trained with gradient-based learning, which is computationally more expensive \cite{fukunaga2004reinforcementLearning}, \cite{campanaro2021cpgActor}.

One approach of dealing with the recurrent nature of CPGs is to treat the CPG controller as part of the environment dynamics. Such approach is referred to as  CPG-Actor-Critic, where the CPG controller, the robot, and the environment are treated as a single dynamic system called CPG-coupled system. The action space of the policy is the CPG parameters. The policy controls the robot by adjusting the CPG parameters of the CPG-coupled system. \cite{cho2019adaptationToEnvironmental},  \cite{nakamura2007reinforcementLearning}, \cite{fukunaga2004reinforcementLearning}. 

Another approach of dealing with the recurrent nature of CPGs is to optimize the parameters of CPGs and MLP network separately. The parameters of the neural network can be optimized with DRL, while the parameters of CPG are optimized using non-gradient based approaches such as evolutionary strategy (ES) \cite{cite:shi2022reinforcementLearning}, genetic algorithm (GA) \cite{wang2021cpgBased}, or a biologically plausible learning rule \cite{wei2018modelOfOperant}, \cite{wei2018bioInspired}. Wang et al. proposed a hierarchical control structure with a high-level neural network and a low-level CPG controller. The high-level network generates latent variables that are fed into the CPGs to regulate their behaviors. The CPG controller is optimized using GA, while the neural network is trained using DRL \cite{wang2021cpgBased}. Shi et al. used ES to train a CPG-based foot trajectory generator. 
A separate neural network is trained using DRL to generate residual joint angles that correct the generated foot trajectory \cite{cite:shi2022reinforcementLearning}.

Campanaro et al. proposed an approach that dealt with the recurrent properties of CPGs by reformulating the CPG into a fully-differentiable stateless network, and thus allowing end-to-end training of CPGs alongside MLP. Campanaro et. al. have only validated the approach on a simple example of a single two degrees-of-freedom (DoF) hopping leg \cite{campanaro2021cpgActor}. We extended upon the idea of reformulating CPG network into a stateless network and proposed our MLP-CPG framework. Furthermore, we have successfully implemented on a 12-DOF simulated quadruped robot.

\section{METHOD}
\subsection{Robot Platform and Simulation setup}
The Jueying quadruped robot is used as the robot platform (Table \ref{tab:robot}). 
Pybullet is chosen as the physics engine for the simulation environment. 

\begin{table}[t]
    \centering
    \caption{Robot specification.}
    \label{tab:robot}
    \begin{tabular}{ p{1.3cm}|p{2.3cm}|p{1.5cm}|p{1.6cm}}
   	\hline
   	     & Joint Range & Peak Torque& Peak Velocity\\ \hline
        Hip roll& (-0.523, 0.523) $rad$ & 75 $Nm$ & 19 $rad/s$ \\
        Hip pitch & (-2.792, 0.349) $rad$ & 75 $Nm$ & 19  $rad/s$\\
        Knee pitch & (0.698, 2.792) $rad$ & 130 $Nm$ & 19 $rad/s$\\
   	\hline
   	Body mass & \multicolumn{3}{c}{42$kg$ }\\
   	Body size & \multicolumn{3}{c}{0.85$m$ x 0.30$m$ x 0.30$m$ (length x width x height)}\\
   	Leg length & \multicolumn{3}{c}{0.33$m$, 0.34$m$ (upper, lower)}\\
   	\hline
    \end{tabular}
    \vspace{-2mm}
\end{table}  

\subsection{CPG Controller}
\begin{figure}[t!]
  \centering
  \includegraphics[width=0.48\textwidth]{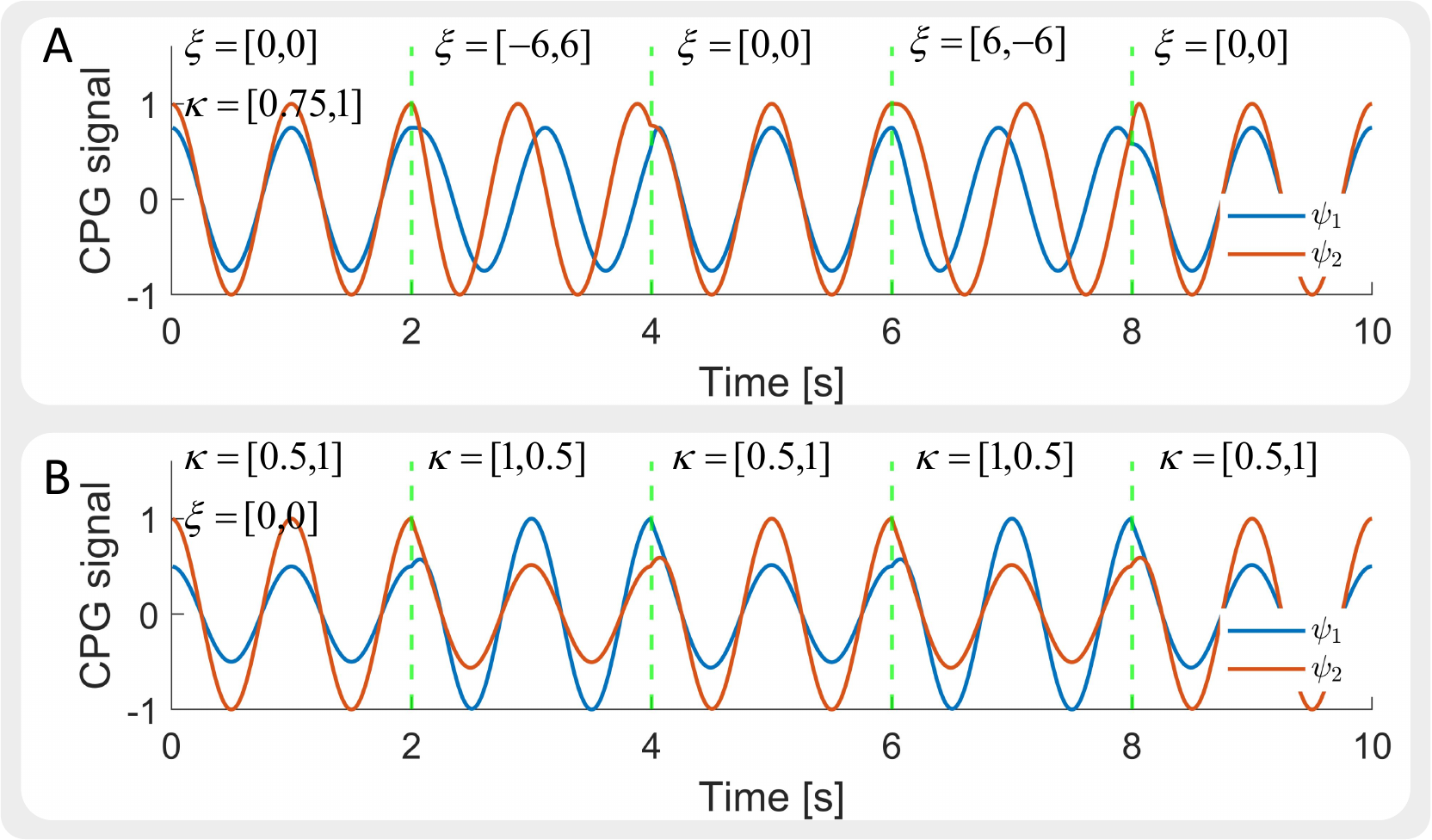}
  \caption{An example consisting of 2 coupled CPG oscillators. The values within $\epsilon_{ij}$, $\gamma$, and $dt$ are set to 6.0, 5.0, and 0.01. The values within $\eta, \chi, o, \phi$ are set to 0. (A) and (B) illustrate how the phase and amplitude of the CPG can be adjusted by feedback component $\xi$ and $\kappa$.}
  \label{CPG}
  \vspace{-6mm}
\end{figure}

The CPGs within this work are modelled using Hopf oscillators \cite{gay2013learningRobot}. 

\begin{equation}
\begin{aligned}
    \dot{\theta}_{i}^{t} &= 2 \pi f^{t}+\sum_{j}\epsilon_{ij}sin(\theta_{j}^{t-1}-\theta_{i}^{t-1}-\phi_{ij})+\xi_{i}^{t}\\
    \dot{r}_{i}^{t} &= \gamma r_{i}^{t-1} \left( (\eta_{i}+\kappa_{i}^{t})^{2} - (r_{i}^{t-1})^{2} \right) \\
   \psi_{i}^{t} &= r_{i}^{t} cos(\theta_{i}^{t}) +\chi_{i}^{t}+o_{i},
   \label{eq:CPG_1}
\end{aligned}    
\end{equation}
where $\theta_{i}^{t}$ and $r_{i}^{t}$ represent the phase and amplitude of the i-th oscillator at timestep $t$, $\dot{\theta}_{i}^{t}$ and $\dot{r}_{i}^{t}$ are their corresponding derivative, $\epsilon_{ij}$ and $\phi_{ij}$ are the coupling weight and phase bias between i-th and j-th oscillator, $\eta_{i}$ is the desired amplitude, $o_{i}$ is a constant offset of the oscillation setpoint, $\gamma$ is a constant that determines the rising time for $r$, $\psi_{i}^{t}$ represents the output signal. Finally, $\kappa_{i}^{t}$, $\chi_{i}^{t}$, $\xi_{i}^{t}$, $f^{t}$ are the feedback components, where $\kappa_{i}^{t}$ adjusts the amplitude, $\xi_{i}^{t}$ adjusts the phase, $\chi_{i}^{t}$ adjusts the oscillation setpoint, $f^{t}$ adjusts the frequency of the oscillator at timestep $t$. 

The phase and amplitude $\theta_{i}^{t}$ and $r_{i}^{t}$ are obtained from their derivative values $\dot{\theta_{i}^{t}}$ and $\dot{r_{i}^{t}}$ using the following equation.

\begin{equation}
\begin{aligned}
    \theta^{t}_{i} &= \theta^{t-1}_{i}+\dot{\theta}^{t}_{i}dt,~
    r^{t}_{i} = r^{t-1}_{i}+\dot{r}^{t}_{i}dt,
    \label{eq:CPG_2}
\end{aligned}    
\end{equation}
where $dt$ is the duration of the timestep.

\begin{figure}[t]
  \centering
  \includegraphics[width=0.48\textwidth]{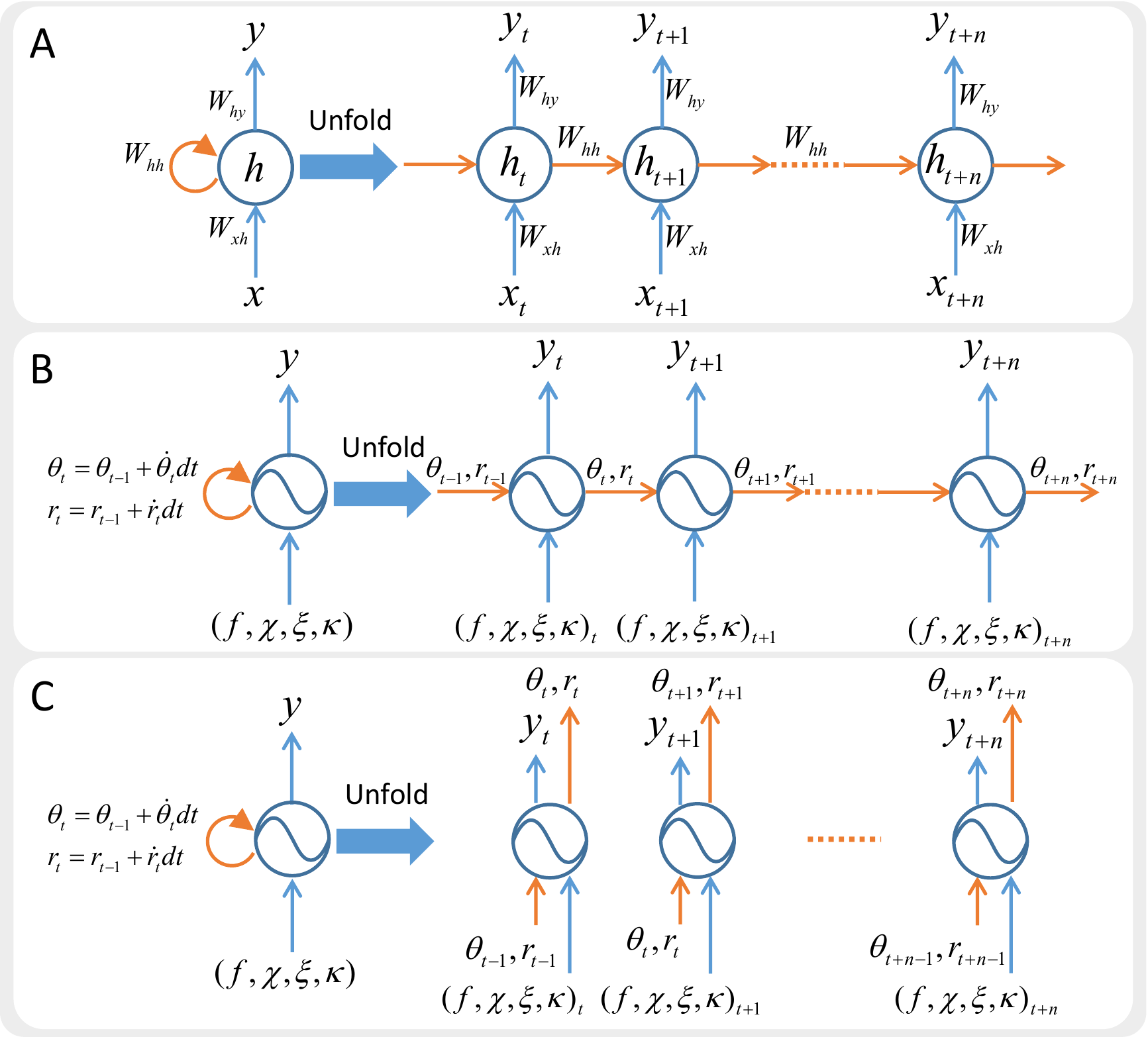}
  \caption{Illustration of RNN and CPGs. (A) Unfolded RNN. (B) Unfolded CPGs. (C) Our reformulation approach transforms CPGs into stateless networks by exposing the hidden states as network inputs and outputs.}
  \label{CPG_RNN}
  \vspace{-6mm}
\end{figure}

\subsection{Reformulating Hidden States in CPGs}

Within Recurrent Neural Networks (RNN), the hidden states of the previous timestep has to be passed to the current timestep and combined with the input of the current timestep to compute the current output. RNN requires BPTT to be trained. 
CPGs can be viewed as a type of RNN, as they require hidden states to propagate information through time. 
The transition between consecutive hidden states of a CPG is determined by eq \eqref{eq:CPG_1} and eq \eqref{eq:CPG_2}.

We dealt with the recurrent nature of CPGs by proposing a reformulation approach that exposes the hidden states through network inputs and outputs. The previous hidden states will be passed as network inputs, and the current hidden states will be passed as network outputs. With our proposed reformulation approach, the CPG network is essentially transformed into a fully-differentiable stateless feedforward network (Fig. \ref{CPG_RNN}C).


In contrast to previous work where the parameters of the CPG controllers and MLP network are optimized with separate algorithms, our reformulated CPG network is differentiable and can be optimized jointly with MLP network within the same DRL paradigm without BPTT \cite{wang2021cpgBased}, \cite{cite:shi2022reinforcementLearning}.

\begin{figure}[t!]
  \centering
  \includegraphics[width=0.48\textwidth]{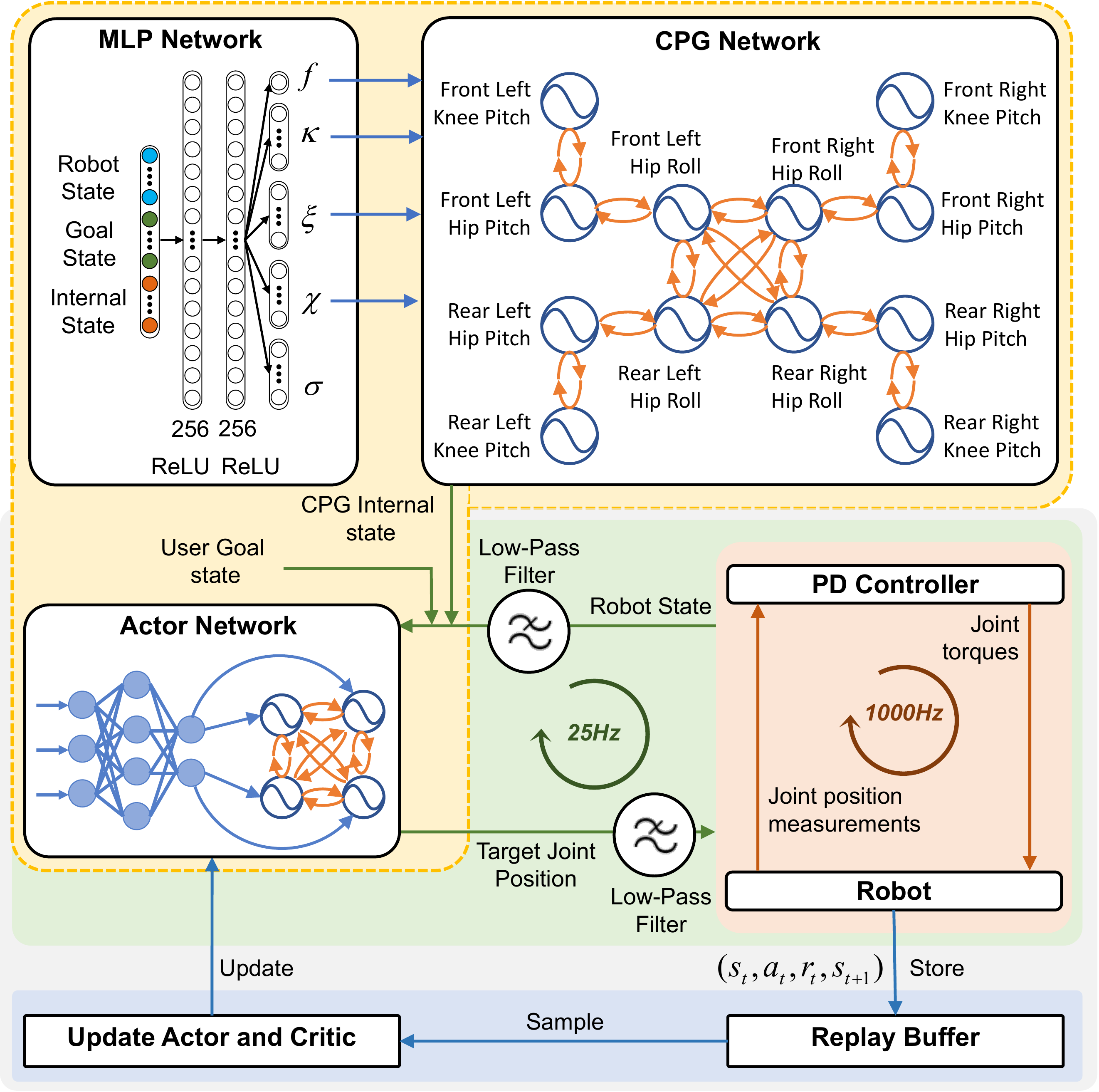}
  \caption{Control framework overview. The MLP-CPG network generates target joint positions at 25Hz. The PD controller functions at 1000Hz and converts the joint angles to joint torques.}
  \label{Control_framework}
  \vspace{-2mm}
\end{figure}

\subsection{Network Architecture}
The MLP-CPG network consists of two separate modules, one rhythmic module responsible for generating rhythmic signals, and another non-linear feedback module responsible for processing sensory feedback (Fig. \ref{Control_framework}).

\subsubsection{Rhythmic module}
The rhythmic module contains 12 CPG neurons, each neuron represents a joint. 
The values of the CPG parameters are shown in Table \ref{tab:CPG}. 

\begin{table}[t]
    \centering
    \caption{CPG parameters.}
    \label{tab:CPG}
    \begin{tabular}{ p{1.5cm}|p{6.0cm}}
   	 \hline
   	    Parameter& Value \\ \hline
   	    $dt$ & 0.04 \\ 
   	    $\gamma$& 12 \\
        $\boldsymbol{\epsilon}$&  Learned via DRL \\ 
   	    $\boldsymbol{\theta}$& Learned via DRL\\
        $\boldsymbol{o}$&  $[\boldsymbol{m}, \boldsymbol{m}, \boldsymbol{m}, \boldsymbol{m}]$, $\boldsymbol{m} = [0, 0.28, -0.1]$\\ 
        $\boldsymbol{\eta}$& $[\boldsymbol{n}, \boldsymbol{n}, \boldsymbol{n}, \boldsymbol{n}]$, $\boldsymbol{n} = [0, 0.8, 0.8]$ \\ 
   	 \hline
    \end{tabular}
    \vspace{-4mm}
\end{table} 

\subsubsection{Feedback module}
The feedback module is an MLP network with 2 hidden layers of size 256 with $\tanh$ activation function. The MLP receives state inputs and regulates the CPG network via feedback components $f$, $\boldsymbol{\kappa}$, $\boldsymbol{\xi}$, $\boldsymbol{\chi}$. 
The MLP also outputs the covariance of the stochastic policy $\boldsymbol{\sigma}$.

The output of the MLP-CPG network $\boldsymbol{\psi}$ is bounded within $(-1,1)$ using $\tanh$, and re-scaled to the corresponding joint limits to be used as the target joint angles.

\subsection{Training Setup}

\subsubsection{Soft Actor Critic}
We select Soft Actor Critic (SAC) as the DRL algorithm for the training of our policy. 
SAC learns a policy by maximizing the expected return and the entropy. The optimization objective can be expressed as follows:

\begin{equation}
    J_{SAC} (\pi) = \sum_{t=0}^{T} \mathbb{E}_
{(s_t ,a_t) \sim \rho_{\pi}} [r(s_t, a_t)
+\alpha H(\pi(\cdot|s_{t}))]
\end{equation}

where $\pi$ is the policy, $\rho_{\pi}$ is the sample distribution, $r$ is the reward, $s_t$ and $a_t$ are the state and action at time step $t$ within the sample distribution, $\alpha$ is the temperature parameter, and $H(\pi)$ is the entropy. The temperature parameter determines the stochasticity of the policy and is tuned automatically during training to balance exploration and exploitation. The hyperparameters can be seen in Table \ref{tab:SAC}


\subsubsection{Generating smooth action}

We introduce temporal and spatial regularization terms to encourage the learning of smoother actions \cite{duburcq2022reactiveStepping}. We only regularize the setpoint feedback component $\chi$ of the network output. 

\begin{equation}
\begin{aligned}
    L_{T} = \left \|\chi(s_{t})-\chi(s_{t+1}) \right \|^{2}_{2}, 
    L_{S} = \left \|\chi(s_{t})-\chi(\hat{s}_{t}) \right \|^{2}_{2}\\
\end{aligned}    
\end{equation}

The spatial regularization loss minimizes the difference between the action generated under observed state $s_{t}$ and perturbed state $\hat{s}_{t} \sim \mathcal{N}(s_{t}, \delta)$
, where $\delta$ is the standard deviation. The temporal regularization loss minimizes the difference between the action under current state $s_{t}$ and the next state $s_{t+1}$.

\subsubsection{Step frequency}

\begin{figure}[t!]
  \centering
  \includegraphics[width=0.48\textwidth]{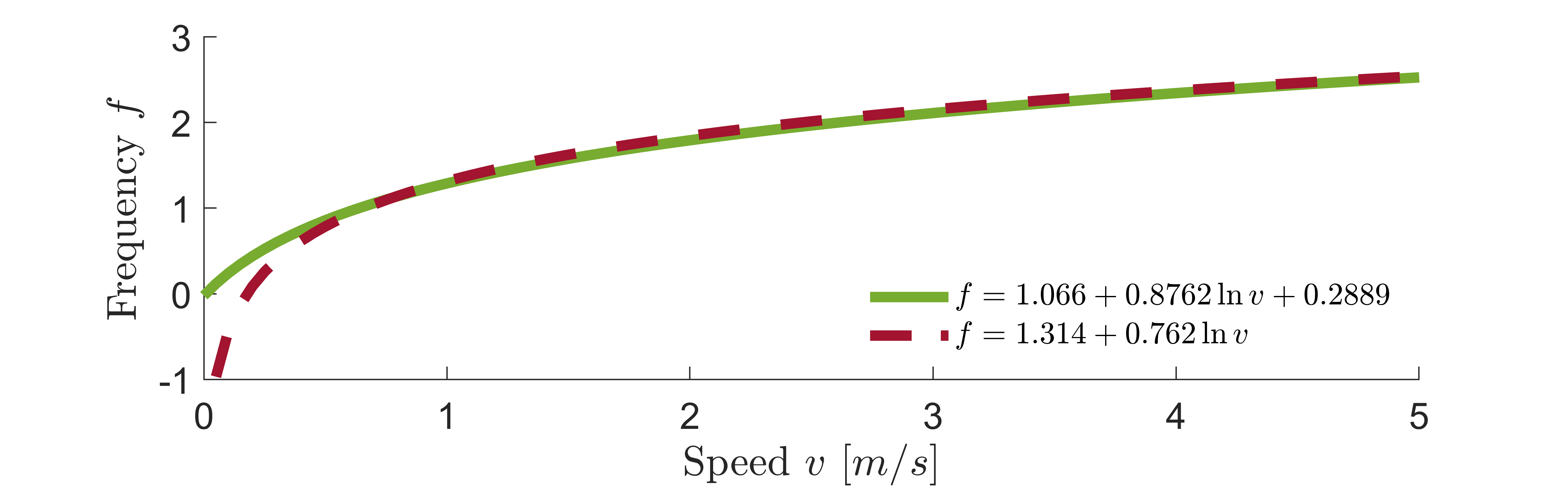}
  \caption{Red dashed curve is obtained from animal data \cite{maes2008steadyLocomotion}. Green solid curve is obtained by fitting a new curve that passes through the origin.}
  \label{Curve}
  \vspace{-2mm}
\end{figure}

\begin{table}[t]
    \centering
    \caption{SAC training hyperparameters.}
    \label{tab:SAC}
    \begin{tabular}{ p{3.0cm}|p{4.5cm}}
   	 \hline
   	 Hyperparameter& Value \\ \hline
   	    Discount factor& 0.95 \\ 
   	    Target network smoothing& 0.999 \\ 
   	    Learning rate& 3e-4 \\
        Weight decay&  1e-6 \\ 
   	    Batch size& 128 \\
        Gradient update steps& 4 \\ 
        Temporal regularization& 1e-3 \\ 
        Spatial regularization& 1e-3 \\ 
        frequency loss& 1e-2 \\
   	 \hline
    \end{tabular}
    \vspace{-4mm}
\end{table}  

Multiple researches have shown that animals adjust their step frequency for different locomotion velocities \cite{maes2008steadyLocomotion}, \cite{granatosky2022strideFrequency}. 
Prior knowledge of the step frequency can be embedded into the controller. Lee et al. manually designed a state machine that switches between different base frequencies depending on user velocity command \cite{lee2020learningQuadrupedal}. 

We provided prior knowledge of step frequency using the curve function $f_{ref} = a+b\ln{v}$ calculated from real animal data, where $f_{ref}$ is the step frequency, $v$ is the locomotion velocity, $a=1.314$ and $b=0.762$ are the parameters \cite{maes2008steadyLocomotion}. 
We fitted a new curve that passes through the origin, as we want zero frequency at zero velocity (Fig. \ref{Curve}): 

\begin{equation}
\begin{aligned}
f_{ref} &= a+b\cdot\ln{(v+c)}, a=1.066, b=0.876, c=0.289 \\
    v &= (\hat{v}_{x}^{2}+\hat{v}_{y}^{2})^{0.5}
    \label{eq:freq}
\end{aligned}    
\end{equation}

\begin{table}[t]
    \centering
    \caption{The basic definitions of the mathematical notation used in the equations for the reward terms}
    \label{tab:nomenclature}
    \begin{tabular}{ p{0.9cm}|p{6.7cm}}
   	 \hline
   	 Notation & Definition\\ \hline
   	    $\boldsymbol{\varphi}_{b}$ & The projection of the gravity vector in the robot base frame\\ 
   	    $h_{b}$ & The robot base height $(z)$ in the world frame \\ 
        $v_{x}$& The forward velocity of the robot base along robot heading \\ 
        $v_{y}$& The lateral velocity of the robot base along robot heading \\ 
        $v_{z}$& The vertical linear velocity of the robot base\\ 
        $\omega_{roll}$& The roll angular velocity of the robot base\\ 
        $\omega_{pitch}$& The pitch angular velocity of the robot base\\
        $\omega_{yaw}$& The yaw angular velocity of the robot base\\
        $\boldsymbol{\tau}$& Torque of joints \\ 
        $\boldsymbol{q}$& Angle of joints \\ 
        $\boldsymbol{\dot{q}}$& Velocity of joints \\ 
        $(\hat{\cdot})$& The desired quantity of placeholder property $(\cdot)$ \\ 
        $\boldsymbol{p}_{f, n}$& The n-th foot horizontal coordinates in the world frame \\ 
        $\boldsymbol{p}_{b}$& The base horizontal coordinates in the world frame \\ 
        $\boldsymbol{p}_{h, n}$& The n-th hip horizontal coordinates in the world frame \\ 
        $h_{f, n}^{w}$& The height of n-th foot in world frame \\ 
        $v_{f, n}^{w}$& The velocity of n-th foot in world frame \\ 
   	 \hline
    \end{tabular}
    \vspace{-2mm}
\end{table}

\begin{table}[t]
    \centering
    \caption{Detailed description of task reward terms.}
    \label{tab:reward}
    \begin{tabular}{ p{2.1cm}|p{5.5cm}}
   	 \hline
   	    Reward term & Symbol \\ \hline
   	    Forward velocity& $\frac{8}{31} \times K(v_{x}, \hat{v}_{x}, -4.6)$ \\ 
   	    Lateral velocity& $\frac{4}{31} \times K(v_{y}, \hat{v}_{y}, -4.6)$ \\ 
   	    Vertical velocity& $\frac{1}{31} \times K(v_{z}, 0, -4.6)$ \\
   	    Roll velocity& $\frac{1}{31} \times K(\omega_{roll}, 0, -1.87)$ \\ 
   	    Pitch velocity& $\frac{1}{31} \times K(\omega_{pitch}, 0, -1.87)$ \\ 
   	    Yaw velocity& $\frac{4}{31} \times K(\omega_{yaw}, \hat{\omega}_{yaw}, -1.87)$ \\
   	    Base orientation& $\frac{2}{31} \times K(\boldsymbol{\varphi}_{b}, [0,0,-1], -2.35)$ \\ 
   	    Base height& $\frac{2}{31} \times K(h_{b}, \hat{h}_{b}, -51.17)$ \\ 
        Joint torque&  $\frac{1}{31} \times K(\boldsymbol{\tau}, 0, -0.001)$ \\ 
   	    Joint velocity& $\frac{1}{31} \times K(\boldsymbol{\dot{q}}, 0, -0.026)$ \\
   	    Ground contact&
        $\frac{1}{31} \times 
        \begin{cases}
        0, &\text{upper body contact with ground}\\
        1, .
        \end{cases}
        $\\   
        Self collision & 
        $\frac{1}{31} \times 
        \begin{cases}
        0, &\text{self collision}\\
        1, .
        \end{cases}
        $\\  
        \\
		Swing \& stance &
		$\frac{2}{31} \times \frac{1}{4} \sum_{4}^{n=1}K((h_{f, n}^{w}-\hat{h}_{f, n}^{w})v_{f, n}^{w}, 0, -460)$ \\
		Body placement &
        $\frac{1}{31} \times K(\frac{1}{4} \sum_{4}^{n=1}(\boldsymbol{p}_{f, n}^{w}), \boldsymbol{p}_{b}^{w}, -51.17)$ \\
        Foot placement& $\frac{1}{31} \times \frac{1}{4} \sum_{4}^{n=1}K(\boldsymbol{p}_{f, n}, \boldsymbol{p}_{h, n}, -51.17)$ \\ 
   	 \hline
    \end{tabular}
    \vspace{-2mm}
\end{table}

where $\hat{v}_{x}$ and $\hat{v}_{y}$ are the target forward and lateral velocity, respectively. 
The step frequency $f$ is bounded within $[0,3]$. We introduce loss function $L_{f}$ to encourage the MLP network to track the reference frequency $f_{ref}$.

\begin{equation}
\begin{aligned}
    L_{f} &= \left \|f(s_{t})-f_{ref} \right \|^{2}_{2}\\
\end{aligned}    
\end{equation}

The frequency loss $L_{f}$, the temporal regularization loss $L_{T}$, and spatial regularization loss $L_{S}$ are added to the loss function of the SAC $L_{SAC} = -J_{SAC}$.
\begin{equation}
\begin{aligned}
    L &= L_{SAC} + \lambda_{T} L_{T} + \lambda_{S} L_{S} + \lambda_{f}L_{f}. 
\end{aligned}    
\end{equation}


\subsection{Reward Design}
Radial basis function (RBF) kernels are used in the reward design. The output range of RBF is bounded within $(0,1)$. The formulation of the RBF kernel is shown as follows:

\begin{equation}
K(x, \widehat{x}, c)=\exp \left(c(\hat{x}-x)^{2}\right)    
\end{equation}

where $x$ is the physical quantity for the evaluation, $\hat{x}$ is the desired value, and $c$ is the parameter that controls the width of the RBF.

\begin{figure*}[t!]
  \centering
  \includegraphics[width=0.98\textwidth]{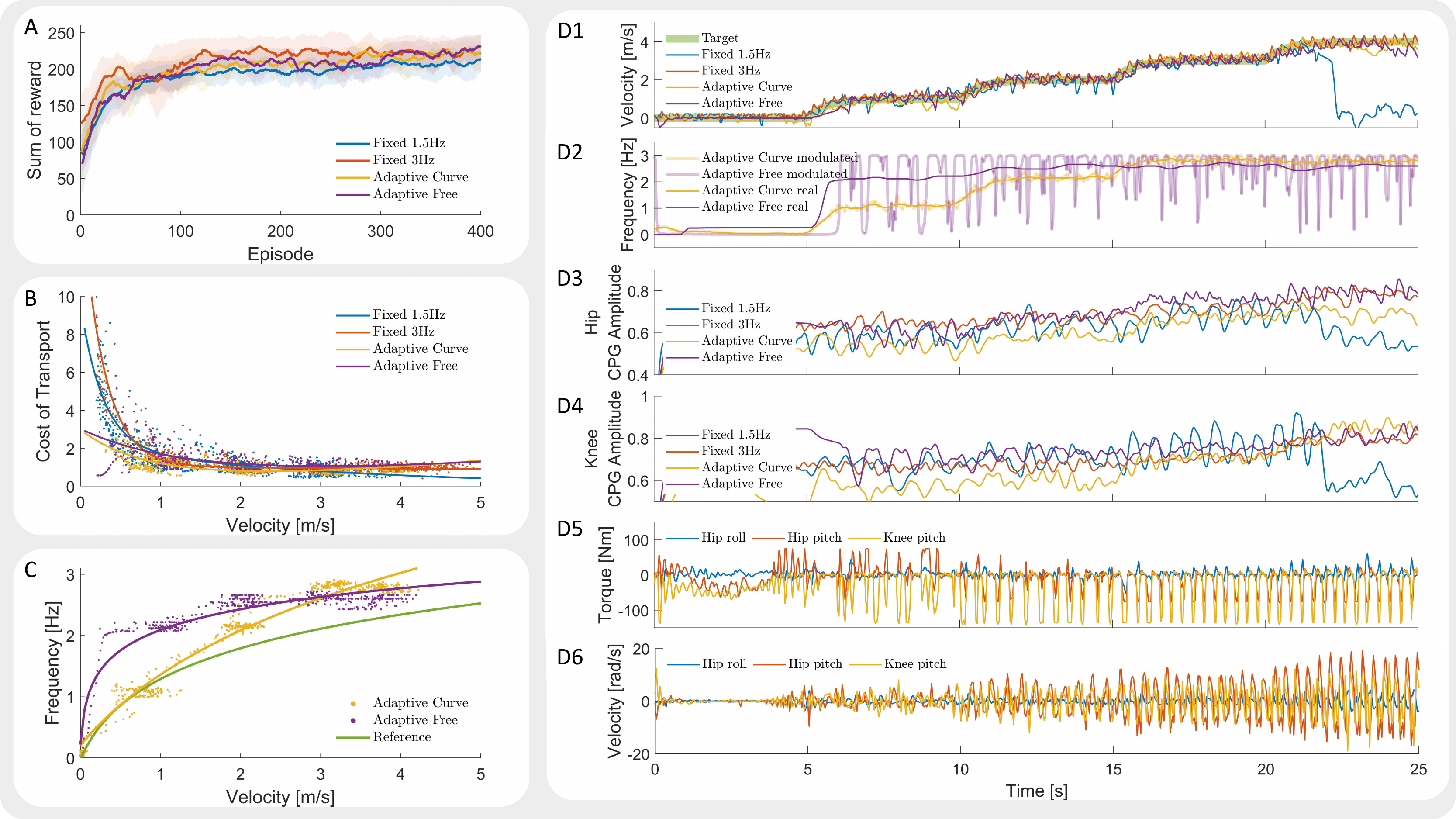}
  \caption{(A) The sum of rewards of \textbf{\textit{Fixed 3.0Hz}}, \textbf{\textit{Adaptive Curve}}, and  \textbf{\textit{Adaptive Free}} converged to similar values, while the sum of reward of \textbf{\textit{Fixed 1.5Hz}} converged to a lower value. (B) Cost of Transport. (C) Correlation between locomotion velocity and step frequency. (D1-D6) The policies are tasked to move forward while tracking a target command velocity. (D1) Target velocity and measured velocity. (D2) Step frequency. Translucent thick lines are the modulated frequency feedback $f$ provided by MLP. Solid thin lines are the real frequency of the gait calculated using time-frequency analysis. (D3 \& D4) Averaged CPG amplitude of the knee and hip joints. (D5 \& D6) Joint torques and velocities from a single leg of policy \textbf{\textit{Adaptive Curve}}.}
  \label{Compare}
  \vspace{-4mm}
\end{figure*}

The reward is the weighted sum of individual reward terms, each governing a different physical aspect (Table \ref{tab:nomenclature} and Table \ref{tab:reward}). 
The Swing and stance foot reward penalizes the movement of the stance foot close to the ground while encouraging the movement of the swing foot far from the ground. The body placement reward encourages the Center of Mass (COM) of the robot to be placed close to the center of the support polygon. The foot placement reward encourages the feet to be placed close beneath the hip joint.

\subsection{State  Space and Action Space}
The state space consists of the observations from the environment $S_{O}$, the internal states of the CPGs $S_{I}$, and the goal states provided by the user $S_{G}$.

The robot observation states consist of base linear velocity in robot heading frame $\boldsymbol{v}_{xyz}$, base angular velocity $\boldsymbol{\omega}_{rpy}$, orientation vector $\boldsymbol{\varphi}_{b}$, joint position $\boldsymbol{q}$, and joint velocity $\boldsymbol{\dot{q}}$. The goal states consist of the target forward velocity $\hat{v}_{x}$, target lateral velocity $\hat{v}_{y}$, and the target yaw turning rate $\hat{\omega}_{z}$. The internal states consist of the amplitude $\boldsymbol{r}$ and phase $\boldsymbol{\theta}$ of the 12 CPG oscillators. A mod operation is used to bound the phase within the range $[0, 2\pi]$. The observation states $S_{O}$ are filtered by a low-pass Butterworth filter with a cut-off frequency of 10$Hz$.

The action space $a_{t}$ contains the target joint angles $\hat{q}$. A low-pass Butterworth filter with a cut-off frequency of 5$Hz$ is applied to the target joint angles to encourage smooth actions. PD controllers are used to calculate joint torques $\tau$ from the target joint angles $\hat{q}$, measured joint angles $q$, and measured joint velocities $\dot{q}$ using the following equation $\tau = K_{p}(\hat{q}-q)+K_{d}(0-\dot{q})$, where $K_{p} = 300$, $k_{d} = 10$. 

\subsection{Exploration Setup}

\subsubsection{Initialization}
The initial forward velocity, lateral velocity, and yaw rate are sampled from $U(-1,5)$, $U(-1,1)$, and $U(-\pi,\pi)$ respectively. The initial joints angles are sampled from $\mathcal{N}(\tilde{q},\frac{\pi}{4})$, where $\tilde{q}$ is the joint angle during nominal standing posture. The CPG amplitude and phase are initialized from $U(0,\frac{\pi}{4})$ and $U(0,2\pi)$, respectively. The target forward velocity, target lateral velocity, and target turning rate are initialized from $U(-1,5)$, $U(-1,1)$, $U(\frac{\pi}{2}, \frac{\pi}{2})$, respectively.

\subsubsection{Early termination}
We terminate and restart the training episode when the robot reaches undesirable fail states. We define the fail state as the state where the body of the robot comes into contact with the ground or when the robot has a large body tilt. We also set a time limit of 10$s$ to terminate the episode early.

\section{RESULTS}
\subsection{Effect of Step Frequency}

Step frequency modulation behaviors have been observed within animals, where the step frequency increases alongside the locomotion velocity. We conduct a comparison study to investigate the effect of step frequency modulation on locomotion performance by comparing the following configurations:

1) \textbf{\textit{Fixed 1.5Hz}}: The intrinsic frequency of the 12 CPGs are fixed to 1.5Hz.

2) \textbf{\textit{Fixed 3.0Hz}}: The intrinsic frequency of the 12 CPGs are fixed to 3Hz.

2) \textbf{\textit{Adaptive Curve}}: Prior knowledge of frequency is provided to encourage the policy to track the reference frequency calculated from eq \eqref{eq:freq}.

3) \textbf{\textit{Adaptive Free}}: No prior knowledge of frequency is provided.

\subsubsection{Locomotion velocity}
Policies from all four configurations are able to perform successful locomotion while tracking user commanded velocity (Fig. \ref{Compare}A). However, the policy from \textbf{\textit{Fixed 1.5Hz}} configuration fails to track 4m/s target velocity, which might be due to the relatively low step frequency (Fig. \ref{Compare}D1).

Figure \ref{Compare}D2 shows the frequency modulation signal $f$ provided by the MLP network. Both policies from \textbf{\textit{Adaptive Curve}} and \textbf{\textit{Adaptive Free}} configurations learned to increase their step frequency to reach higher locomotion velocity (Fig. \ref{Compare}C). While frequency modulation behavior naturally emerges without providing prior knowledge, providing prior knowledge of step frequency through a loss function leads to a smoother frequency modulation behavior.

\begin{figure*}[t!]
  \centering
  \includegraphics[width=0.98\textwidth]{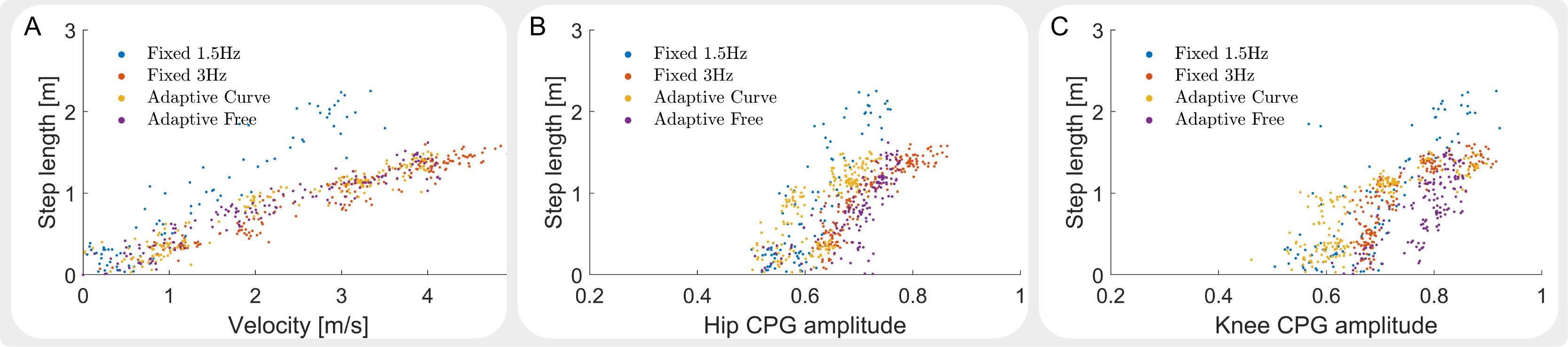}
  \caption{(A) Correlation between locomotion velocity and step length. Step length increases accordingly as the velocity increases. (B) and (C) Correlation between CPG oscillation amplitude of the hip and knee. Step length increases as the CPG oscillation amplitude increases}
  \label{Step_length}
  \vspace{-4mm}
\end{figure*}

\subsubsection{Energy efficiency}
We utilize Cost of Transport (CoT) to measure the energy efficiency of the policies under different locomotion velocities. After comparing the policies from four configurations, we observed that policies with fixed frequency exhibit significantly higher CoT within the low-velocity region compared to policies capable of self-modulating frequency (Fig. \ref{Compare}B). This is due to unnecessarily high-frequency stepping.

\subsubsection{Step length}
Animals adjust their step length to regulate their locomotion velocity \cite{granatosky2022strideFrequency}. We observed similar emergent behavior of step length modulation among all four policies. All policies have learned to increase the step length to reach higher locomotion velocities (Fig. \ref{Step_length}A). Policies modulate the step length via the adjustments of the amplitude of the joint CPGs. The amplitude of the hip and knee CPG oscillators within all policies changes accordingly with locomotion velocity (Fig. \ref{Compare}D3 \& D4). The change in amplitude will have a direct impact on the step length as can be seen in Fig. \ref{Step_length}B and Fig. \ref{Step_length}C, where larger CPG oscillation amplitude results in a larger step length. The step length for policy \textbf{\textit{Fixed 1.5Hz}} is higher compared to the other three policies to compensate for the relatively lower step frequency.

\begin{figure*}[t]
  \centering
  \includegraphics[width=0.98\textwidth]{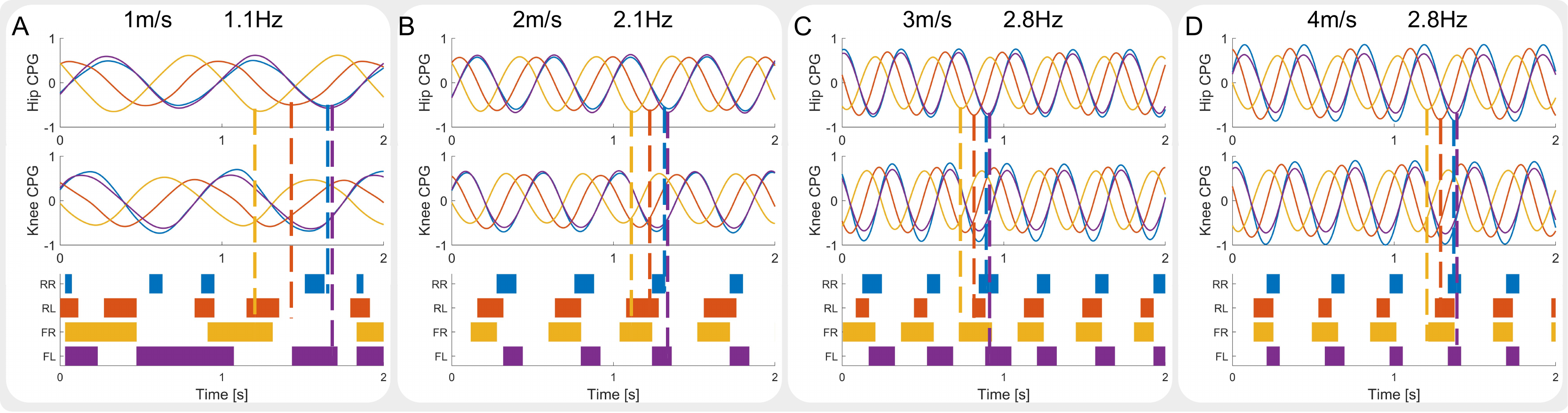}
  \caption{Joint phase, step frequency, and foot contact pattern of the \textbf{\textit{Adaptive Curve}} policy under different locomotion velocities. (A) 1.1$Hz$ step frequency under 1$m/s$. (B) 2.1$Hz$ step frequency under 2$m/s$. (C) 2.8$Hz$  step frequency under 3$m/s$. (D) 2.8$Hz$ step frequency under 4$m/s$. FL: front left, FR: front right, RL: rear left, RR: rear right.}
  \label{Phase}
  \vspace{-3mm}
\end{figure*}


\begin{figure*}[t!]
  \centering
  \includegraphics[width=0.98\textwidth]{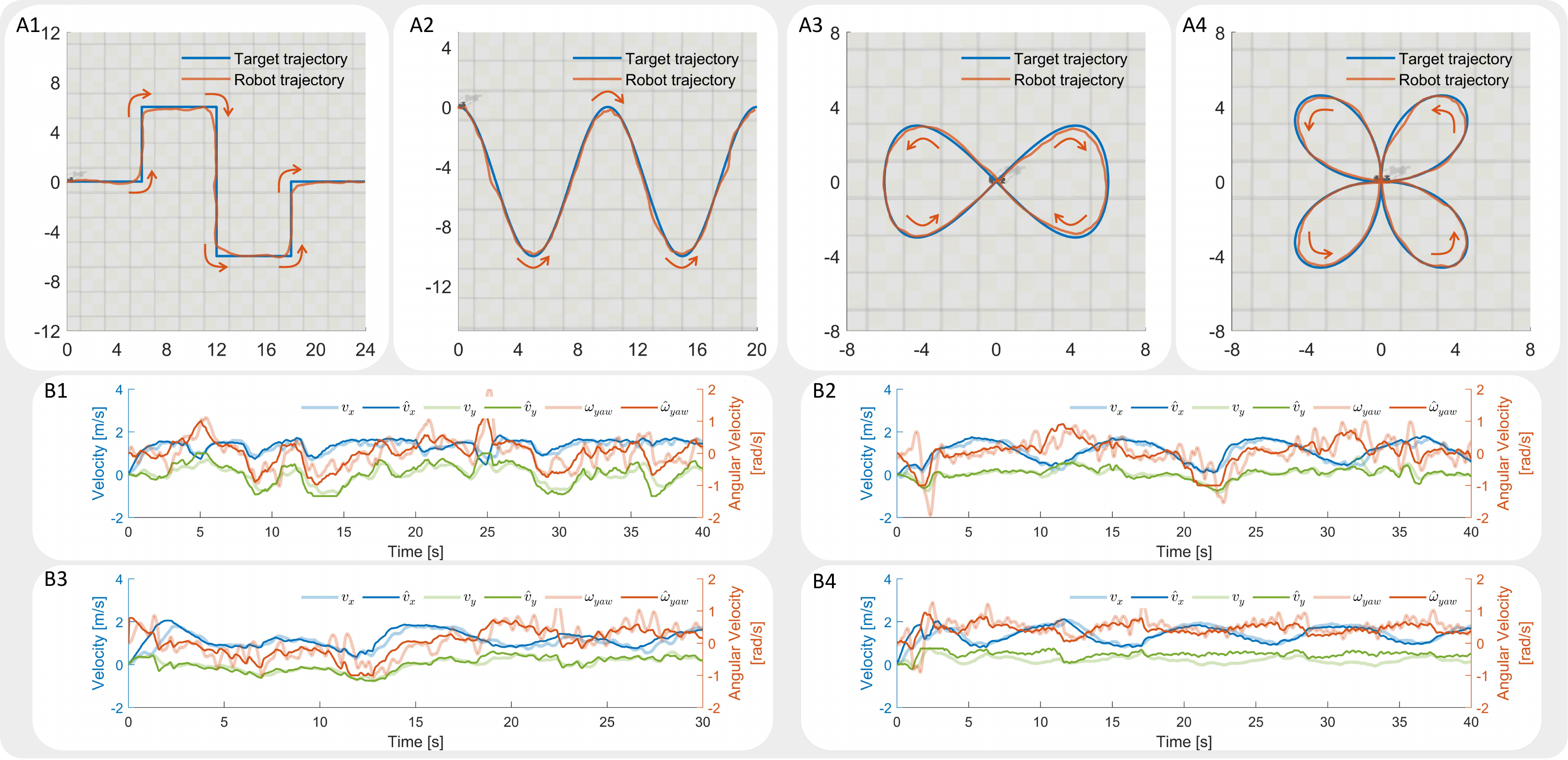}
  \caption{(A1-A4) Trajectory following performance of policy \textbf{\textit{Adaptive Curve}}. (A1) Square Wave. (A2) Cosine Wave. (A3) Eight Curve. (A4) Four-leaved Clover Curve. (B1-B4) The velocity tracking performance while following trajectory A1-A4.}
  \label{Trajectory_following}
  \vspace{-4mm}
\end{figure*}

\subsection{Interpretability of MLP-CPG Architecture}
A side benefit of our MLP-CPG network architecture is the additional interpretability compared to vanilla neural networks. The frequency, amplitude, and phase of the CPG neurons are parameters that can be directly retrieved from the MLP-CPG network. As can be seen from Fig. \ref{Step_length} and Fig. \ref{Phase}, the frequency, amplitude, and phase of the CPGs correlate with step frequency, step length, and foot contact pattern, which are three important aspects of legged locomotion. By analyzing the values of the frequency, amplitude, and phase of the CPG neurons, we can easily interpret the locomotion characteristics of the robot.

\subsection{Performance Benchmark}



\subsubsection{Target following}

We design 4 sets of trajectories with varying difficulties for the robot to follow (Fig. \ref{Trajectory_following}). The robot is tasked to follow the trajectory by tracking a moving target. 
The relative position of the moving target is translated into velocity command for the robot. The policy from \textbf{\textit{Adaptive Curve}} is evaluated.

Given the target position and robot base position, we calculate the velocity commands as follows:
\begin{equation}
\begin{aligned}
    \hat{v}_{x} = - x_{target}^{base}, \hat{v}_{y} = - y_{target}^{base}, 
    \hat{\omega}_{yaw} = -\arctan(\frac{y_{target}^{base}}{x_{target}^{base}}),
\end{aligned}
\end{equation}
where $x_{target}^{base}$ is the x coordinate of the target in robot base frame, and $y_{target}^{base}$ is the y coordinate of the target in robot base frame. The velocity targets are bounded within $[-1, 4]~m/s$, $[-0.75, 0.75]~m/s$, $[-1.0, 1.0]~rad/s$ for forward velocity, lateral velocity, and yaw angular rate. The learned \textbf{\textit{Adaptive Curve}} policy is able to follow all four trajectories.

\subsection{Robustness Tests}
\begin{figure}[t]
  \centering
  \includegraphics[width=0.48\textwidth]{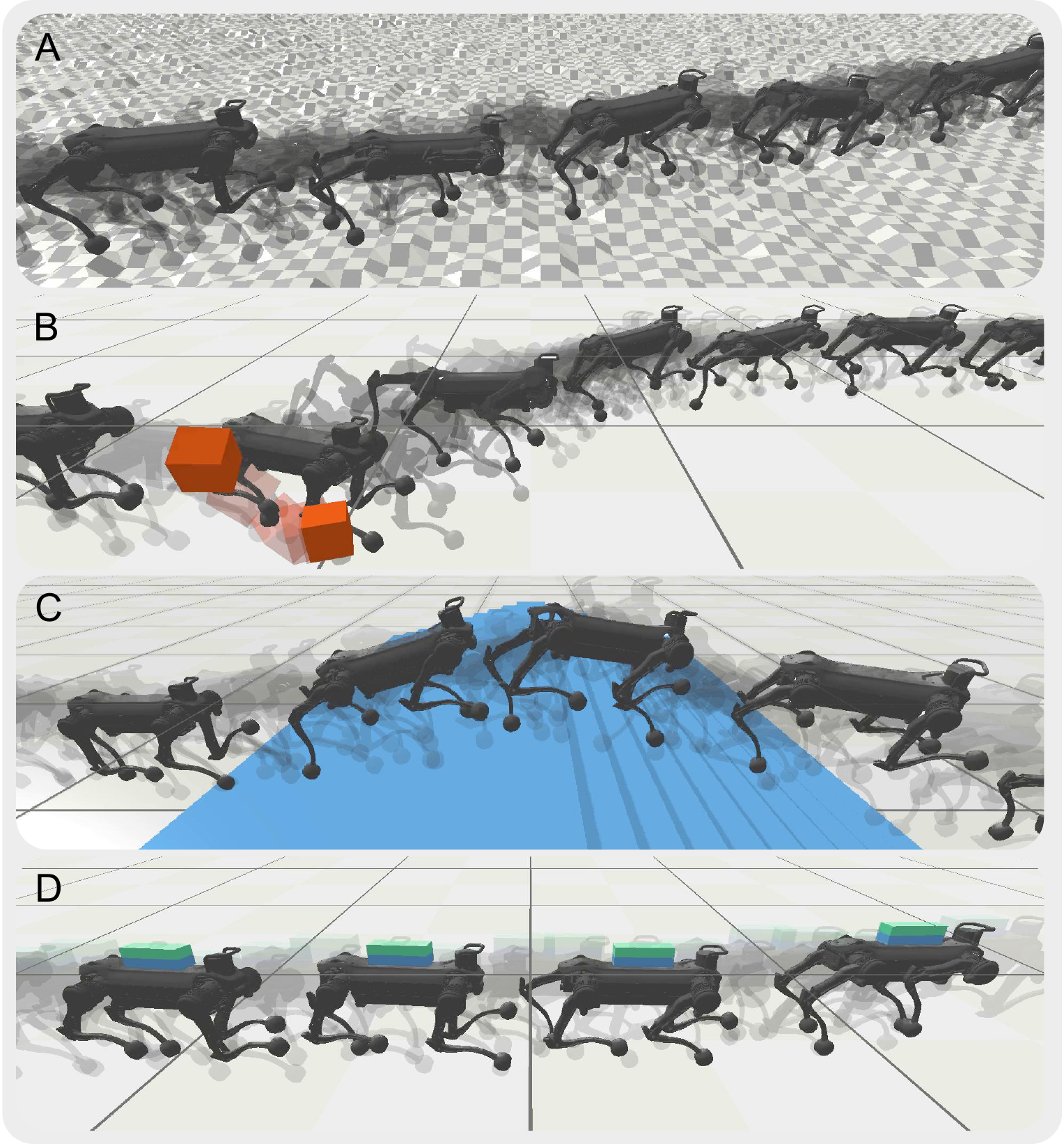}
  \caption{Overlaid snapshots of the \textbf{\textit{Adaptive Curve}} policy. (A) Uneven terrain. (B) External disturbance. (C) Stairs. (D) External load.}
  \label{Robustness}
  \vspace{-6mm}
\end{figure}

We design four sets of test scenarios to evaluate the robustness of the learned policies (Fig. \ref{Robustness}). 1) Blind traversal over Uneven terrain. The uneven terrain consists of small bumps with height ranging from 0$m$ to 0.1$m$. 2) Persistent perturbation. External perturbation forces are applied to the robot every 4s through the impact of small cubes with mass of 5$kg$. 3) Blind traversal over stairs. The stair consists of steps  with height of 0.05m. 4) Carrying external load. A load with mass of 20$kg$ (50\% of the robot's own mass) is placed on the robot.

Policies from all four training configurations are able to traverse over uneven terrain and stairs, carry external load, and resist external perturbations (See accompanying video). The policies exhibit highly dynamic motions while reacting to external uncertainties in the environment. Note that the policies are trained on flat ground and encountered no perturbations during the training phase.

\section{Discussion and Future work}

While our work demonstrated robust locomotion in simulation, it has not yet been implemented on a real robot. For future work, we plan to transfer the policy to a real robot system. Additionally, high-speed locomotion on real robot systems is challenging. Even though our policy is capable of achieving velocities up to 4m/s in simulation while satisfying motor constraints, having the joint torques and velocities constantly reaching the motor limits will inevitably cause a lot of stress and also overheat the motors. Mechanical stress and motor overheating need to be considered during sim-to-real transfer.

This paper has only focused on achieving rhythmic motor task of locomotion for quadruped robots. For future work, we will investigate the feasibility of using our proposed MLP-CPG network to achieve multi-tasking of both rhythmic motor tasks such as locomotion and non-rhythmic motor tasks such as fall recovery.


\section{CONCLUSION}
In this paper, we proposed a bio-inspired network architecture called MLP-CPG that utilizes the intrinsic rhythmicity of CPGs to generate rhythmic motor patterns for locomotion without relying on external phase input. 
We proposed a reformulation approach that transforms CPGs into stateless networks by exposing the internal states. With our approach, the parameters of CPGs can be optimized jointly with the MLP network using DRL.

Our MLP-CPG has the advantage of interpretability. Locomotion characteristics such as step frequency, step length, gait pattern of the robot can be inferred by analyzing the values of the frequency, amplitude, and phase of the CPGs, 

Our MLP-CPG network demonstrates emergent behaviors of step frequency and step length modulation. We observed that the learned policies increase the step frequency and step length accordingly as the locomotion velocity increases, demonstrating that the policy can actively alter locomotion patterns. The learned policy is not only capable of successfully performing the task of target following, but also capable of robustly traversing over uneven terrain and stairs, carrying external load, and resisting external perturbations.


\addtolength{\textheight}{-12cm}   






\bibliographystyle{IEEEtran}
\bibliography{IEEEabrv,reference}

\begin{thebibliography}{10}
\providecommand{\url}[1]{#1}
\csname url@rmstyle\endcsname
\providecommand{\newblock}{\relax}
\providecommand{\bibinfo}[2]{#2}
\providecommand\BIBentrySTDinterwordspacing{\spaceskip=0pt\relax}
\providecommand\BIBentryALTinterwordstretchfactor{4}
\providecommand\BIBentryALTinterwordspacing{\spaceskip=\fontdimen2\font plus
\BIBentryALTinterwordstretchfactor\fontdimen3\font minus
  \fontdimen4\font\relax}
\providecommand\BIBforeignlanguage[2]{{%
\expandafter\ifx\csname l@#1\endcsname\relax
\typeout{** WARNING: IEEEtran.bst: No hyphenation pattern has been}%
\typeout{** loaded for the language `#1'. Using the pattern for}%
\typeout{** the default language instead.}%
\else
\language=\csname l@#1\endcsname
\fi
#2}}

\bibitem{yu2013surveyOn}
J.~Yu, M.~Tan, J.~Chen, and J.~Zhang, ``A survey on cpg-inspired control models
  and system implementation,'' \emph{IEEE transactions on neural networks and
  learning systems}, vol.~25, no.~3, pp. 441--456, 2013.

\bibitem{ijspeert2008centralPattern}
A.~J. Ijspeert, ``Central pattern generators for locomotion control in animals
  and robots: a review,'' \emph{Neural networks}, vol.~21, no.~4, pp. 642--653,
  2008.

\bibitem{yang2020multiExpert}
C.~Yang, K.~Yuan, Q.~Zhu, W.~Yu, and Z.~Li, ``Multi-expert learning of adaptive
  legged locomotion,'' \emph{Science Robotics}, vol.~5, no.~49, p. eabb2174,
  2020.

\bibitem{lee2020learningQuadrupedal}
J.~Lee, J.~Hwangbo, L.~Wellhausen, V.~Koltun, and M.~Hutter, ``Learning
  quadrupedal locomotion over challenging terrain,'' \emph{Science robotics},
  vol.~5, no.~47, p. eabc5986, 2020.

\bibitem{shao2021learningFree}
Y.~Shao, Y.~Jin, X.~Liu, W.~He, H.~Wang, and W.~Yang, ``Learning free gait
  transition for quadruped robots via phase-guided controller,'' \emph{IEEE
  Robotics and Automation Letters}, 2021.

\bibitem{sheng2022bioInspired}
J.~Sheng, Y.~Chen, X.~Fang, W.~Zhang, R.~Song, Y.~Zheng, and Y.~Li,
  ``Bio-inspired rhythmic locomotion for quadruped robots,'' \emph{IEEE
  Robotics and Automation Letters}, 2022.

\bibitem{yang2022fastAnd}
Y.~Yang, T.~Zhang, E.~Coumans, J.~Tan, and B.~Boots, ``Fast and efficient
  locomotion via learned gait transitions,'' in \emph{Conference on Robot
  Learning}.\hskip 1em plus 0.5em minus 0.4em\relax PMLR, 2022, pp. 773--783.

\bibitem{cho2019adaptationToEnvironmental}
Y.~Cho, S.~Manzoor, and Y.~Choi, ``Adaptation to environmental change using
  reinforcement learning for robotic salamander,'' \emph{Intelligent Service
  Robotics}, vol.~12, no.~3, pp. 209--218, 2019.

\bibitem{nakamura2007reinforcementLearning}
Y.~Nakamura, T.~Mori, M.-a. Sato, and S.~Ishii, ``Reinforcement learning for a
  biped robot based on a cpg-actor-critic method,'' \emph{Neural networks},
  vol.~20, no.~6, pp. 723--735, 2007.

\bibitem{fukunaga2004reinforcementLearning}
S.~Fukunaga, Y.~Nakamura, K.~Aso, and S.~Ishii, ``Reinforcement learning for a
  snake-like robot controlled by a central pattern generator,'' in \emph{IEEE
  Conference on Robotics, Automation and Mechatronics, 2004.}, vol.~2.\hskip
  1em plus 0.5em minus 0.4em\relax IEEE, 2004, pp. 909--914.

\bibitem{aoi2017adaptiveControl}
S.~Aoi, P.~Manoonpong, Y.~Ambe, F.~Matsuno, and F.~W{\"o}rg{\"o}tter,
  ``Adaptive control strategies for interlimb coordination in legged robots: a
  review,'' \emph{Frontiers in neurorobotics}, vol.~11, p.~39, 2017.

\bibitem{thor2019fastOnline}
M.~Thor and P.~Manoonpong, ``A fast online frequency adaptation mechanism for
  cpg-based robot motion control,'' \emph{IEEE Robotics and Automation
  Letters}, vol.~4, no.~4, pp. 3324--3331, 2019.

\bibitem{endo2005experimentalStudies}
G.~Endo, J.~Nakanishi, J.~Morimoto, and G.~Cheng, ``Experimental studies of a
  neural oscillator for biped locomotion with qrio,'' in \emph{Proceedings of
  the 2005 IEEE international conference on robotics and automation}.\hskip 1em
  plus 0.5em minus 0.4em\relax IEEE, 2005, pp. 596--602.

\bibitem{gay2013learningRobot}
S.~Gay, J.~Santos-Victor, and A.~Ijspeert, ``Learning robot gait stability
  using neural networks as sensory feedback function for central pattern
  generators,'' in \emph{2013 IEEE/RSJ international conference on intelligent
  robots and systems}.\hskip 1em plus 0.5em minus 0.4em\relax Ieee, 2013, pp.
  194--201.

\bibitem{campanaro2021cpgActor}
L.~Campanaro, S.~Gangapurwala, D.~D. Martini, W.~Merkt, and I.~Havoutis,
  ``Cpg-actor: Reinforcement learning for central pattern generators,'' in
  \emph{Annual Conference Towards Autonomous Robotic Systems}.\hskip 1em plus
  0.5em minus 0.4em\relax Springer, 2021, pp. 25--35.

\bibitem{cite:shi2022reinforcementLearning}
H.~Shi, B.~Zhou, H.~Zeng, F.~Wang, Y.~Dong, J.~Li, K.~Wang, H.~Tian, and
  M.~Q.-H. Meng, ``Reinforcement learning with evolutionary trajectory
  generator: A general approach for quadrupedal locomotion,'' \emph{IEEE
  Robotics and Automation Letters}, vol.~7, no.~2, pp. 3085--3092, 2022.

\bibitem{wang2021cpgBased}
J.~Wang, C.~Hu, and Y.~Zhu, ``Cpg-based hierarchical locomotion control for
  modular quadrupedal robots using deep reinforcement learning,'' \emph{IEEE
  Robotics and Automation Letters}, vol.~6, no.~4, pp. 7193--7200, 2021.

\bibitem{wei2018modelOfOperant}
T.~Wei and B.~Webb, ``A model of operant learning based on chaotically varying
  synaptic strength,'' \emph{Neural Networks}, vol. 108, pp. 114--127, 2018.

\bibitem{wei2018bioInspired}
------, ``A bio-inspired reinforcement learning rule to optimise dynamical
  neural networks for robot control,'' in \emph{2018 IEEE/RSJ International
  Conference on Intelligent Robots and Systems (IROS)}.\hskip 1em plus 0.5em
  minus 0.4em\relax IEEE, 2018, pp. 556--561.

\bibitem{duburcq2022reactiveStepping}
A.~Duburcq, F.~Schramm, G.~Bo{\'e}ris, N.~Bredeche, and Y.~Chevaleyre,
  ``Reactive stepping for humanoid robots using reinforcement learning:
  Application to standing push recovery on the exoskeleton atalante,''
  \emph{arXiv preprint arXiv:2203.01148}, 2022.

\bibitem{maes2008steadyLocomotion}
L.~D. Maes, M.~Herbin, R.~Hackert, V.~L. Bels, and A.~Abourachid, ``Steady
  locomotion in dogs: temporal and associated spatial coordination patterns and
  the effect of speed,'' \emph{Journal of Experimental Biology}, vol. 211,
  no.~1, pp. 138--149, 2008.

\bibitem{granatosky2022strideFrequency}
M.~C. Granatosky and E.~J. McElroy, ``Stride frequency or length? a
  phylogenetic approach to understand how animals regulate locomotor speed,''
  \emph{Journal of Experimental Biology}, vol. 225, no. Suppl\_1, p. jeb243231,
  2022.

\end{thebibliography}

\end{document}